\documentclass[sigconf]{acmart}
\usepackage{float}
\usepackage{multirow}
\usepackage{lscape}
\usepackage{subfig}

\AtBeginDocument{%
  \providecommand\BibTeX{{%
    \normalfont B\kern-0.5em{\scshape i\kern-0.25em b}\kern-0.8em\TeX}}}
    
\copyrightyear{2021}
\acmYear{2021}
\setcopyright{acmlicensed}\acmConference[ARIC'21]{4th ACM SIGSPATIAL International Workshop on Advances in Resilient and Intelligent Cities }{November 2, 2021}{Beijing, China}
\acmBooktitle{4th ACM SIGSPATIAL International Workshop on Advances in Resilient and Intelligent Cities (ARIC'21), November 2, 2021, Beijing, China}
\acmPrice{15.00}
\acmDOI{10.1145/3486626.3493434}
\acmISBN{978-1-4503-9116-0/21/11}

\begin{document}

\title{Transfer Learning Approach to Bicycle-sharing Systems' Station Location Planning using OpenStreetMap Data}

\author{Kamil Raczycki}
\email{raczyckikamil@gmail.com}
\orcid{0000-0002-3715-4869}
\affiliation{
  \institution{Department of Computational Intelligence\\Wrocław University of Science and Technology}
  \city{Wrocław}
  \country{Poland}}

\author{Piotr Szymański}
\email{piotr.szymanski@pwr.edu.pl}
\orcid{0000-0002-7733-3239}
\affiliation{
  \institution{Department of Computational Intelligence\\Wrocław University of Science and Technology}
  \city{Wrocław}
  \country{Poland}}

\renewcommand{\shortauthors}{Raczycki and Szymański}
\renewcommand{\shorttitle}{Transfer learning approach to bicycle-sharing systems' station location planning using OpenStreetMap data}

\begin{abstract}
  Bicycle-sharing systems (BSS) have become a daily reality for many citizens of larger, wealthier cities in developed regions. However, planning the layout of bicycle-sharing stations usually requires expensive data gathering, surveying travel behavior and trip modelling followed by station layout optimization. Many smaller cities and towns, especially in developing areas, may have difficulty financing such projects. Planning a BSS also takes a considerable amount of time. Yet as the pandemic has shown us, municipalities will face the need to adapt rapidly to mobility shifts, which include citizens leaving public transport for bicycles. Laying out a bike sharing system quickly will become critical in addressing the increase in bike demand. This paper addresses the problem of cost and time in BSS layout design and proposes a new solution to streamline and facilitate the process of such planning by using spatial embedding methods. Based only on publicly available data from OpenStreetMap, and station layouts from 34 cities in Europe, a method has been developed to divide cities into micro-regions using the Uber H3 discrete global grid system and to indicate regions where it is worth placing a station based on existing systems in different cities using transfer learning. The result of the work is a mechanism to support planners in their decision making when planning a station layout with a choice of reference cities.
\end{abstract}





\begin{teaserfigure}
  \includegraphics[width=\textwidth]{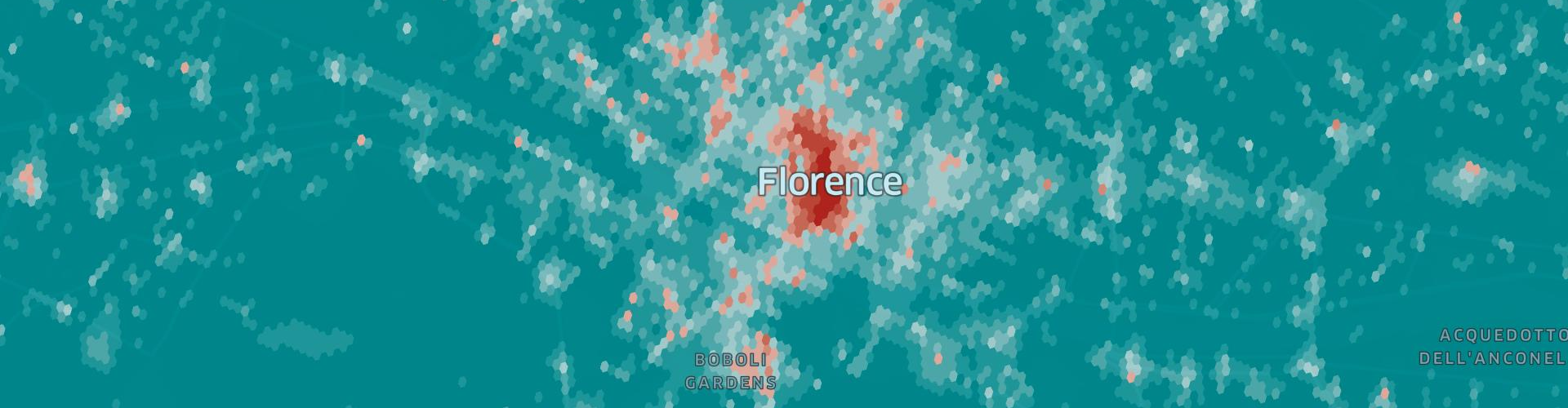}
  \caption{Bicycle-sharing station probability predictions in microregions for the city of Florence, Italy.}
  \label{fig:teaser}
\end{teaserfigure}

\maketitle

\section{Introduction}
In the history of the last 50 years, bicycle-sharing systems (BSSs) have evolved and adapted to growing cities. For many users, the shared bike is a comfortable mean of coming back home from an urban evening with friends or completing a short errand. For some, in the most reliable of systems, it may be a daily mean of transport. However, designing a BSS layout is not a cheap endeavor. It usually requires gathering data about population density, surveying citizens about their travel patterns, building a trip model to assess bike travel demand. Once such analyses are performed, one has enough data to use one of the well-established optimization models to layout bicycle stations across the city. This process of BSS design is often the case when public institutions want to create a subsidized BSS with a private partner. In this form, the process of planning a BSS layout, especially for a public tender, is expensive and lengthy. A significant amount of money is therefore redirected from the public subsidy for the BSS to the preparation and modelling steps, impacting negatively the number of bikes and stations available in the final systems, as well as delaying its deployment.

For many regions, even the basic urban data is not publicly available or is limited. The data is also stored in different formats, which results in the need to process it or adapt existing methods and models. Such tasks generate costs and often take a long time. One of the tasks related to urban data is planning the layout of bicycle-sharing stations. The currently available methods for BSS layout optimization focus on manual selection and processing of traffic and station neighborhood features using a limited number of POI types. 

The complexity of BSS layout design has become an even stronger issue in the unstable times of the climate catastrophe. The COVID-19 pandemic caused major shifts in mobility patterns. As the safety of public transport was still uncertain, citizen mobility shifted to individual means of transport. This phenomenon forced cities to take quick infrastructural decisions to ensure better safety of new bike traveler. Due to a rise of bicycle demand, accessibility to bicycles became limited. Many cities benefited from existing bicycle-sharing systems, which facilitated the new covid mobility trends. Cities without an existing BSS were facing problems in a resilient response, as deploying a new BSS quickly was not possible due to the complexity and cost of the layout design process. Many smaller cities and towns, especially in developing areas, find it difficult to finance such projects even in the economically secure times. However, most of them do have access to an urban planner with a reasonable intuition in local characteristics. 

In this paper, we address the problem of cost and time in BSS layout design and proposes a new solution to streamline and facilitate the process of such planning by using spatial embedding methods. We hope to provide urban planning offices in cities with a data-driven support to their intuitions by providing a transfer-learning based model for BSS planning. Based only on publicly available data from OpenStreetMap, and station layouts from 34 cities in Europe, a method has been developed to divide cities into micro-regions using the Uber H3 discrete global grid system and to indicate regions where it is worth placing a station based on existing systems in different cities using transfer learning. The result of the work is a mechanism to support planners in their decision-making when planning a station layout with a choice of reference cities.

The region embedding method proposed in this paper is intended to allow machine learning capabilities to be used to a greater extent than currently available methods. It will use publicly available data from OpenStreetMap, thanks to which it will be possible to apply the method in every city which has spatial data entered into this system. The method will focus on embedding a city region arbitrarily divided into regular polygons and predict whether a station should be located in a particular region or not. In the context of the task of planning the position of city bicycle sharing stations, it is supposed to propose an initial layout of stations in a city without any special preparation of data for a specific city, which will allow planners later to elaborate the plan in more detail. We publish our code, models and reports in a public github repository\footnote{\url{https://github.com/pwr-inf/Transfer-learning-approach-to-bicycle-sharing-systems-station-location-planning-using-OpenStreetMap/}}

\section{Related works}

Researchers on the subject categorise existing bicycle sharing systems into five generations \cite{CHEN_2018}:
\begin{enumerate}
    \item free bikes available to the public (Amsterdam, Netherlands, 1965),
    \item bikes available for a cash deposit (Copenhagen, Denmark, 1991),
    \item bikes with locking stations unlocked by magnetic card (Portsmouth University, UK, 1996),
    \item bikes rented using a mobile application linked to an ITS system and providing real-time data,
    \item dockless bikes that can be picked up and returned anywhere in a service area. 
\end{enumerate}

The bicycle-sharing systems discussed in the context of this study are mainly of the 4th generation. For Madrid, \citet{GarcaPalomares2012} describe a GIS-based method that consists of four steps: estimating potential user demand, finding station positions based on demand, collecting characteristics of proposed stations and finally analysing these stations in terms of accessibility to potential destinations. \citet{Liu_2015} developed a sophisticated method to optimise the existing station layout in New York. In addition to bicycle mobility data, publicly available taxi traffic data and category information for more than 27,000 POIs were used. \citet{Park_2017} focused on the city's administrative district of Gangnam-gu in Seoul, which had no bicycle-sharing stations and it's goal was to determine potential bike station positions, using the trajectories of taxis passing through the study district; demand points from a selected set of points: metro stations, shopping malls, parks, and residences; and solve two modes of the Location-Allocation model \cite{iiasa1781}: the minimum impedance (p-median) and the maximum location coverage problem (MCLP). \citet{Cintrano_2020} proposes the use of metaheuristics to optimise the layout of stations in Malaga, Spain; also attempting to minimise impedance (i.e. distances between residents and bike stations) in a p-median problem. The following methods were investigated: genetic algorithm, iterated local search, particle swarm optimisation, variable neighbourhood search and simulated annealing. \cite{Yang_2020} tackle the problem for Wuhan, China; and propose to use the aspect of temporal demand variability with spatial information when planning station layout. Based on very accurate GPS data on bicycle use in agglomeration, the work builds a spatial-temporal bicycle demand cube. In Istanbul, Turkey - \citet{Guler_2021}, propose the use of the best worst-case method (BWM) based on several variables related to the demand for station presence. A more open-data oriented approach, leveraging demographics, POI and social media mining was presented by \cite{10.1145/2750858.2804291}. Crowdsourced information was also used in optimization-based approaches by \cite{10.1145/3209582.3209583,gao2018collaborative}. The approach closest to the one presented in this paper was proposed by \cite{wang2020leveraging} - who recommend bike sites based on OSM map tile embedding as imagery using convolutional neural networks and POIs extracted from the amenity tag.

The topic of planning the layout of bicycle-sharing stations is important and is discussed quite widely in the literature, \cite{nath2019modelling} is a good introduction to the topic. Most works focus either on optimizing the existing layout for a selected city/region or on planning the layout from scratch. What is almost universally common is that the task is being solved for a single geographical area using a wide variety of local data. Unfortunately, these methods are highly complex, require significant work to be applied - such as mobility modelling \cite{yang2019mobility} -  in a different city and use nonpublic data that are made available locally by a municipal organization. Results also often depend ont the knowledge of a domain expert. Not all municipalities can afford such analysis or they simply do not collect the necessary data, so the methods described in the review cannot be applied. Our approach aims to provide a new path to solving bicycle-sharing station localization problems by learning from multiple cities' already existing systems.

\section{Study area and data}

For the analysis, 34 European cities were selected which have implemented bicycle-sharing systems. All these systems have at least 100 stations in their infrastructure. 

The BikeShareMap website and the Nextbike API were used to obtain data on the rental systems, and the Overpass API from OpenStreetMap was used to download spatial data. Information about the bike stations was removed from the OSM features dataset. Objects were of different OSM types (node, way, relation) and were grouped into 20 categories: 
\begin{enumerate}
    \item \texttt{aerialway} - air transport elements like gondolas and cable cars;
    \item \texttt{airports} - air transport infrastructure;
    \item \texttt{buildings} - any buildings not included in other categories (like offices);
    \item \texttt{culture\_and\_entertainment} - cultural and entertainment facilities;
    \item \texttt{education} - education facilities from nurseries to university campuses;
    \item \texttt{emergency} - emergency facilities such as EDs, defibrillators and medical helicopter landing pads;
    \item \texttt{finances} - banks, exchange offices and ATMs;
    \item \texttt{healthcare} - all medical buildings and pharmacies;
    \item \texttt{historic} - historical sites such as ruins and historical monuments;
    \item \texttt{leisure} - leisure facilities;
    \item \texttt{other} - public buildings: orphanages, cemeteries, embassies, post offices, prisons, places of worship, police and fire stations, shopping centres, courts of law;
    \item \texttt{roads\_bike} - roads where bicycles can ride;
    \item \texttt{roads\_drive} - roads where motorised vehicles can ride;
    \item \texttt{roads\_walk} - roads where pedestrians can walk;
    \item \texttt{shops} - any shops;
    \item \texttt{sport} - sporting venues;
    \item \texttt{sustenance} - restaurants, cafes, bars, pubs;
    \item \texttt{tourism} - tourist facilities: hotels, lodging places, tourist attractions, museums, zoos, amusement parks;
    \item \texttt{transportation} - car parks, public transport stops, railway stations, also bicycle sharing stations (they were removed from the dataset so that the model would not learn on them);
    \item \texttt{water} - bodies of water, seas, beaches, rivers, canals.
\end{enumerate}

\begin{figure*}[t]
  \centering
  \includegraphics[width=\textwidth]{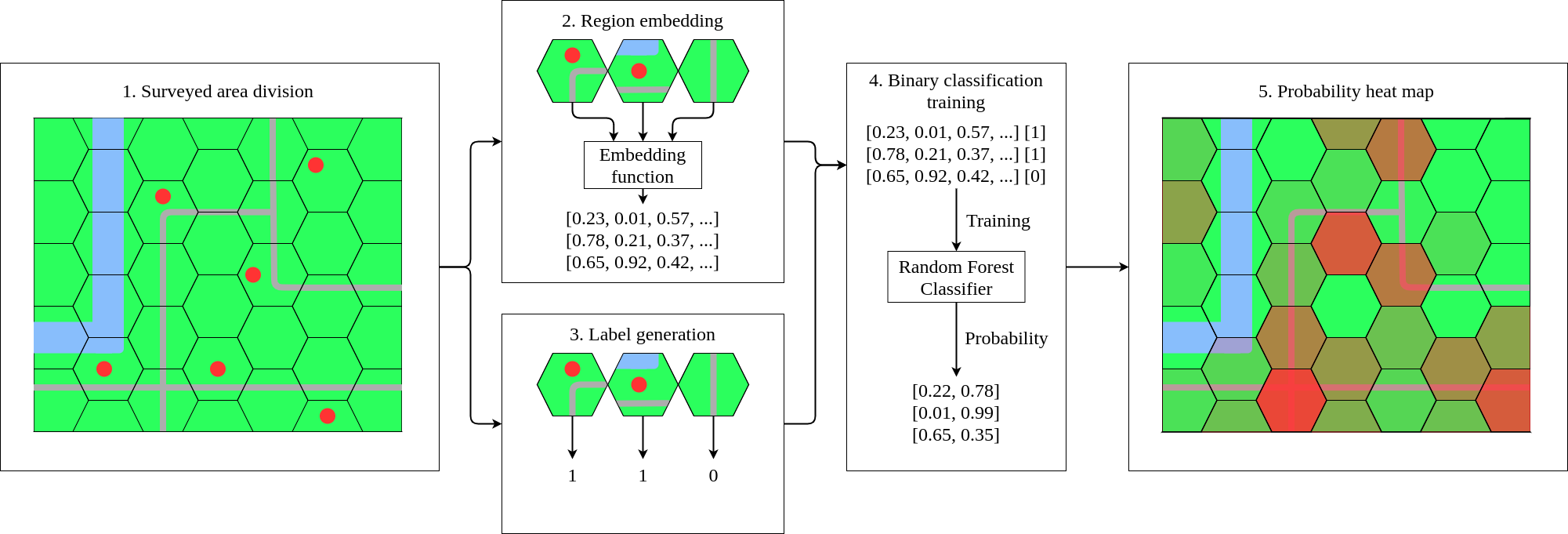}
  \caption{Methodological framework of our prediction model.}
\label{fig:framework}
\end{figure*}
The surveyed cities were divided into regions using the Uber H3 library at resolutions 9, 10, and 11.

The resulting dataset contained 10,360 bike stations and 2,787,408 OSM spatial objects. Surveyed regions in all cities had a total area of almost 11,000 km$^2$.

\section{Methodology}

\subsection{Methodological framework}

In this section, we present the basic framework of our method as shown in Figure \ref{fig:framework}. The operation of the method consists of 5 steps described below.

Firstly, the areas are divided into regions. This is done using the Uber H3 Discrete Global Grid system. For each region, the OpenStreetMap objects that reside in that region are assigned.

The second step is to embed the objects inside each region into a numerical vector using an embedding function. Region neighbourhood can also be taken into account when generating feature vectors. At the end, a city-scoped min-max normalisation is applied per column.

In parallel, a partitioning is made into regions containing stations and those without stations to build binary labels for learning.

The fourth step is to learn the classifier based on the obtained feature vectors and labels in the binary classification task.

Finally, using the learned classifier, a prediction is made and the obtained probabilities are converted into a heat map.

\subsection{Generating region embeddings}

To obtain the embedding vector for a region, 2 main methods have been proposed: a baseline method involving per-category counting, and a method that analyses the shapes of objects per category. 

The first method, similarly to word counting in sentences in natural language processing, counts how many objects from each of the 20 defined categories occur within the region. 

To extend the first method and to take into account the shape of the surveyed objects, a second method is proposed which analyses the polygon areas and the lengths of the road lines inside the region. 4 categories are treated specially: in case of \texttt{water} only the areas occupied by water inside the region are counted and in case of \texttt{roads\_bike}, \texttt{roads\_drive} and \texttt{roads\_walk} the sum of lengths of roads inside the region is counted. For the other 16 categories, the areas of objects inside the region are counted separately if they are marked as area and the number of occurrences if the object is marked as a point. The final dimension of the vector is 36 ($16 \cdot 2 + 4 \cdot 1$).

To take advantage of the additional data granularity resulting from OSM tags such as \texttt{amenity}, \texttt{building}, \texttt{shop}, \texttt{sport} and others, it was decided to expand the second method. For most of the 16 categories (except water and roads), it has been decided to choose a tag that uniquely identifies the second hierarchy of importance of the object. Based on this second hierarchy, a vector of dimension 5702 is generated. However, based on different regions, this number can vary because it depends on all different values that users can add. This results in a very sparse vector. In order not to take this vector directly into the learning, an autoencoder was used to reduce its dimensionality to 300.

Additionally, a set of the most popular tags was selected to filter out any rarely used tags added by users. This resulted in a vector of dimension 888, which was also reduced using an autoencoder to multiple values from 20 to 500.

\subsection{Embedding region neighbourhood}

\begin{figure}[h]
  \centering
  \includegraphics[width=\linewidth]{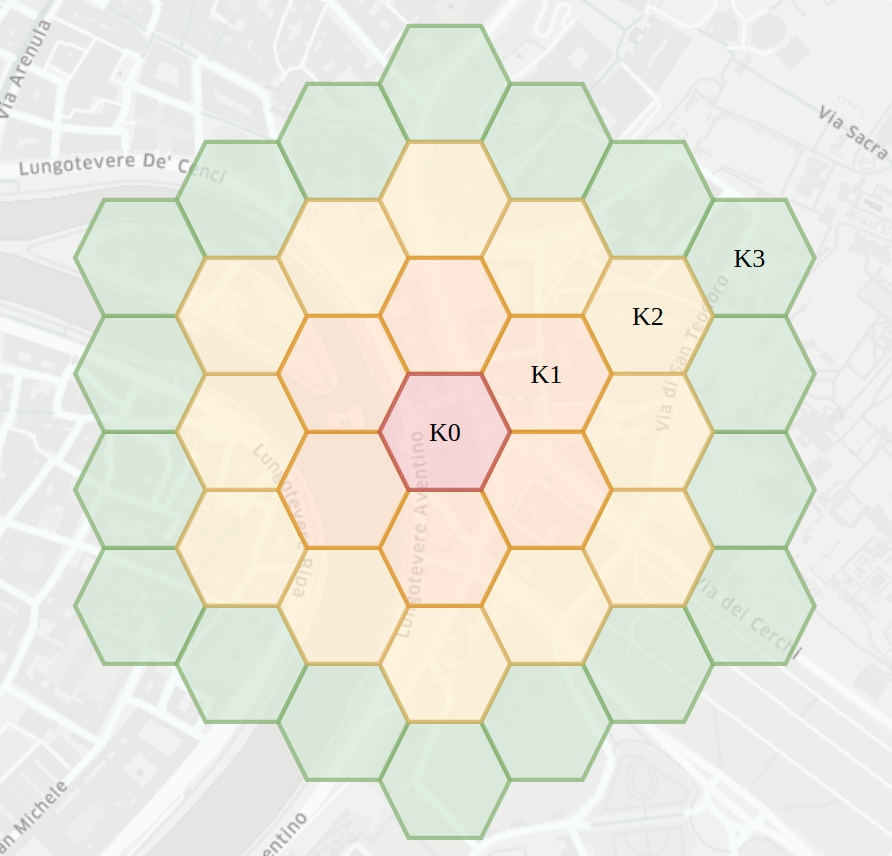}
  \caption[Example of different sizes of regions neighbourhood.]{Example of different sizes of regions neighbourhood. Each neighbourhood ring has been coloured and labelled correspondingly.}
\label{fig:neighbourhood_example}
\end{figure}

To provide the classifier with more context for making predictions, it was decided that the embedding vector of a particular region would also contain information about the neighbourhood of that region. The neighbourhood of a region includes hexagons adjacent to the hexagon under study. In this paper, the neighbourhood size parameter will be investigated, which translates to the number of rings of consecutive hexagons surrounding the region under study. An illustrative diagram showing the neighbourhoods can be found in Figure \ref{fig:neighbourhood_example}. To facilitate the method's operation, it was decided that the embedding vectors generated for all regions included in one neighbourhood ring will be averaged and only the vectors of particular neighbourhood rings will be treated separately. Four methods of combining embedding vectors have been proposed and are described below.

\paragraph*{Concatenation}

A basic method that involves combining vectors of consecutive neighbourhoods with each other. On the one hand, this preserves all information, but on the other hand, it increases the computational complexity as the size of the neighbourhood considered increases. The final vector is of dimension $N\cdot (K+1)$ where $N$ is the dimension of the single region vector and $K$ is the size of the neighbourhood.

\paragraph*{Averaging}

The second base method takes all vectors for consecutive neighbourhoods and averages them together with the vector generated for the region under study. This ensures that the vector has a constant dimension regardless of the size of the neighbourhood, but with more rings of neighbours taken into account, the noise may increase as all vectors are averaged equally.

\paragraph*{Diminishing averaging}

Extending the previous method by adding a weighting aspect. The vector for the study region is assigned a weight of 1 and each subsequent neighbourhood is assigned a weight of $1 / (k+1)$ where $k$ is the distance of the ring of neighbours from the centre. Finally, the vectors are averaged taking into account the assigned weights, so that further neighbourhoods have less influence on the change in values.

\paragraph*{Diminishing averaging squared}

Another variant of weighted averaging, but this time the assigned weight for successive neighbourhoods equals $1 / (k+1)^2$. This was proposed to test which version of weight assignment might have a better impact on the predictions made by the proposed method.

\subsection{The class imbalance problem}

The resulting dataset contains only (depending on the resolution tested) between $2.99\cdot 10^{-3}$ and $8.95\cdot 10^{-2}$ regions that contain stations. If the model was taught on the whole set, it could turn out that there are big problems in the prediction of station occurrence, because more than 99\% of regions are examples without stations. For this purpose, limiting the set of negative examples will be used during learning and the effect of the ratio between positive and negative classes on the quality of prediction and discrimination of regions will be investigated. When building the learning set for a given city, all regions containing stations will be used and an appropriate number of regions that do not contain stations will be drawn at random. Values between 1 and 5 will be examined and compared to see how they affect the method results and the meaningfulness of the results obtained as a heat map.

\section{Results}

\subsection{Model evaluation indices}

In this paper, 4 classical metrics were used to measure the quality of the model performance in the station occurrence prediction task: Accuracy, F1 score, Precision, and Recall.

\subsection{Impact of the region neighbourhood embedding method on the prediction quality}

\begin{figure}[h]
    \centering
    \includegraphics[width=\linewidth]{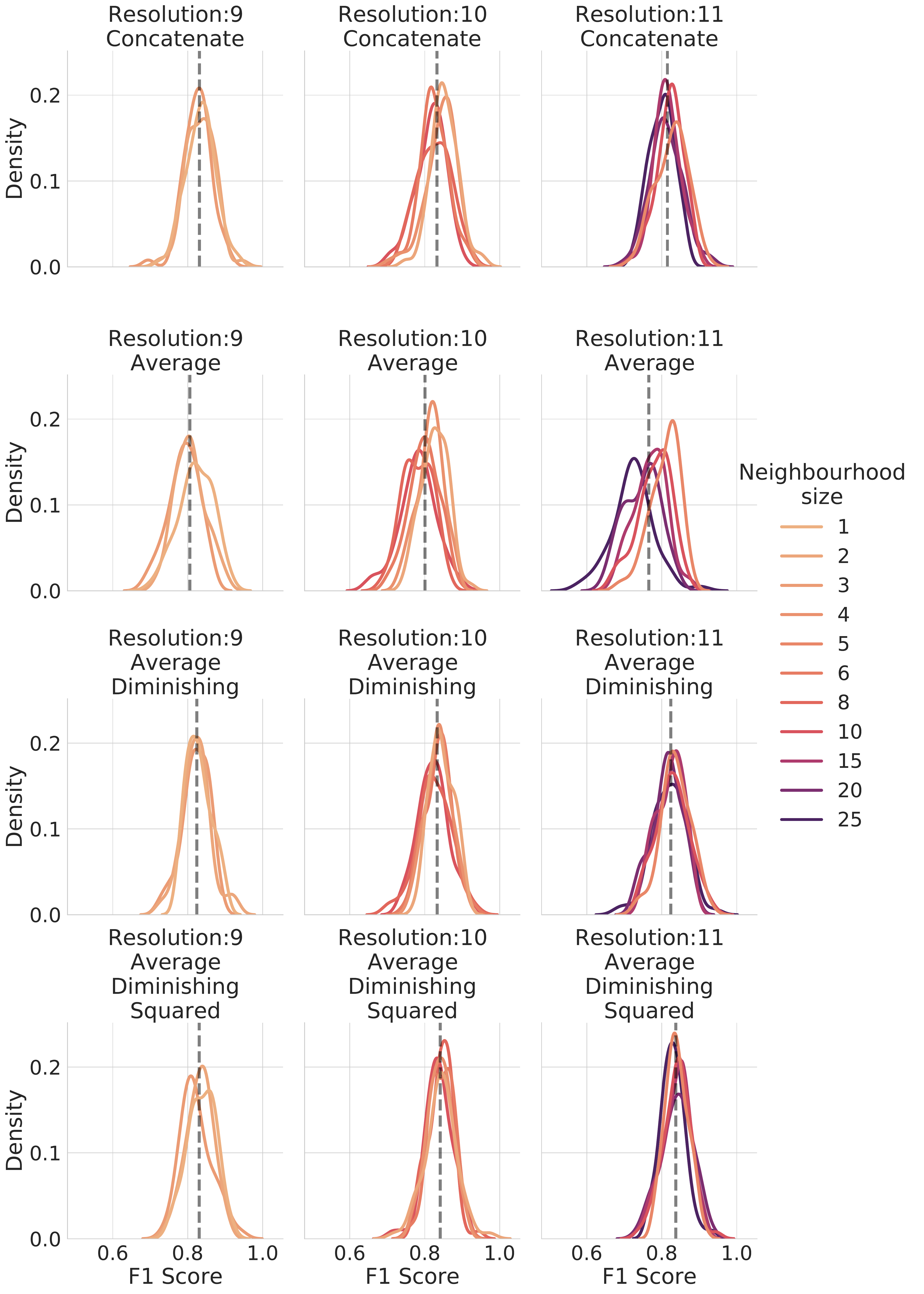}
    \caption[Distribution of prediction results for different neighbourhood embedding methods and hex resolutions]%
    {Distribution of prediction results for different neighbourhood embedding methods and hex resolutions. The vertical dashed lines indicate the average value in a particular chart.}
    \label{chart:neighbour_embedding_methods}
\end{figure}

The first methods' hyperparameter investigated is the choice of the neighbourhood embedding method. Its performance directly affects the quality and speed of the classifier. The distribution of the results obtained can be seen in Figure \ref{chart:neighbour_embedding_methods} and the averaged results can be found in Table \ref{tab:comparing_neighbourhood_emb}.

\begin{table}[]
\centering
\caption[Comparison of different neighbourhood embedding methods]%
{Comparison of different neighbourhood embedding methods. Results for the cities of Poznań, Warsaw, and Wrocław. Values represent the average of all grouped results for different parameters. The best values in each row were highlighted.}
\label{tab:comparing_neighbourhood_emb}
\resizebox{\linewidth}{!}{%
\begin{tabular}{@{}cc|cccc@{}}
\toprule
\begin{tabular}[c]{@{}c@{}}Hex\\ resolution\end{tabular} &
  Metric &
  Concatenate &
  Average &
  \begin{tabular}[c]{@{}c@{}}Average\\ Diminishing\end{tabular} &
  \begin{tabular}[c]{@{}c@{}}Average\\ Diminishing\\ Squared\end{tabular} \\ \midrule
\multirow{2}{*}{9}  & Accuracy & 0.824 & 0.795 & 0.821 & \textbf{0.826} \\
                    & F1 Score & \textbf{0.831} & 0.802 & 0.826 & \textbf{0.831} \\ \midrule
\multirow{2}{*}{10} & Accuracy & 0.828 & 0.791 & 0.826 & \textbf{0.836} \\
                    & F1 Score & 0.835 & 0.797 & 0.832 & \textbf{0.841} \\ \midrule
\multirow{2}{*}{11} & Accuracy & 0.805 & 0.761 & 0.815 & \textbf{0.830}  \\
                    & F1 Score & 0.813 & 0.768 & 0.820 & \textbf{0.834} \\ \bottomrule
\end{tabular}%
}
\end{table}

The collected results indicate that the best proposed method is the squared diminishing averaging of the embedding vectors of successive neighbourhoods. For resolution 9, the concatenation of vectors was able to equal the prediction quality, but for higher resolutions, it deteriorated more and more. It can also be seen that the last method produces the most stable results regardless of the size of the neighbourhood under study, where the averaging alone method significantly degraded the prediction quality as the neighbourhood size increased.

\subsection{Impact of the region embedding method on the prediction quality}

Another hyperparameter investigated is the single region embedding method. The methods differ in complexity and generate vectors of different lengths, which may have a direct impact on the results obtained. The first method, Category Counting, is the simplest of the proposed ones and it simply counts the number of objects present inside the region per 20 defined categories. The second method, Shape Analysis, takes into account the appearance aspect of the object by counting the area and length. It is used in 3 modes: analysis per category, analysis per all tags available in OSM, and analysis per selected OSM tags. The first two methods, Category Counting and Shape Analysis per category are tested using both a random forest classifier and a neural network. The other methods were passed through an autoencoder with different final vector lengths and then used in a neural network. The results obtained are listed in Table \ref{tab:comparing_region_emb}.

The best results were obtained with the simplest method (Category counting) in combination with the random forest classifier.

\subsection{Impact of the class imbalance ratio on the prediction quality}

In two previous investigations, to maintain a balance in learning between regions with and without stations, all micro-regions with stations were used in the learning set and an equal number of microregions without stations were sampled. The ratio between the classes was 1:1. Unfortunately, in this approach only a small subset of the regions from the entire city coverage is selected during the draw, thus a large part of the city area is missed. The effect is exacerbated at higher resolutions as the size of the regions decreases. Therefore, this step will investigate the effect of the imbalance ratio on the obtained results. The goal is to use the highest possible ratio to cover as much city size as possible in the learning, without significant loss of F1 Score measure prediction quality. In addition to the baseline random forest classifier, two values of the class balance parameter will be investigated: balanced and balanced per sample, to see if these parameters will produce better results than the baseline classifier. 

Intuitively, as the class imbalance increases, the accuracy increases and the F1 score decreases. The quite high precision value indicates that a high percentage of all regions predicted to contain stations actually contain them. In contrast, the rapidly decreasing recall value indicates that the model has difficulty predicting all regions containing stations and marks some of them as not containing stations. This may be due to the situation when the direct neighbourhood of the region that contains stations is used in the learning - then there is a probability that 6 neighbouring regions have very similar embedding vector as the central region and there is an overrepresentation of similar regions as not containing stations in the training dataset.

\begin{figure}[h]
    \centering
    \includegraphics[width=0.8\linewidth]{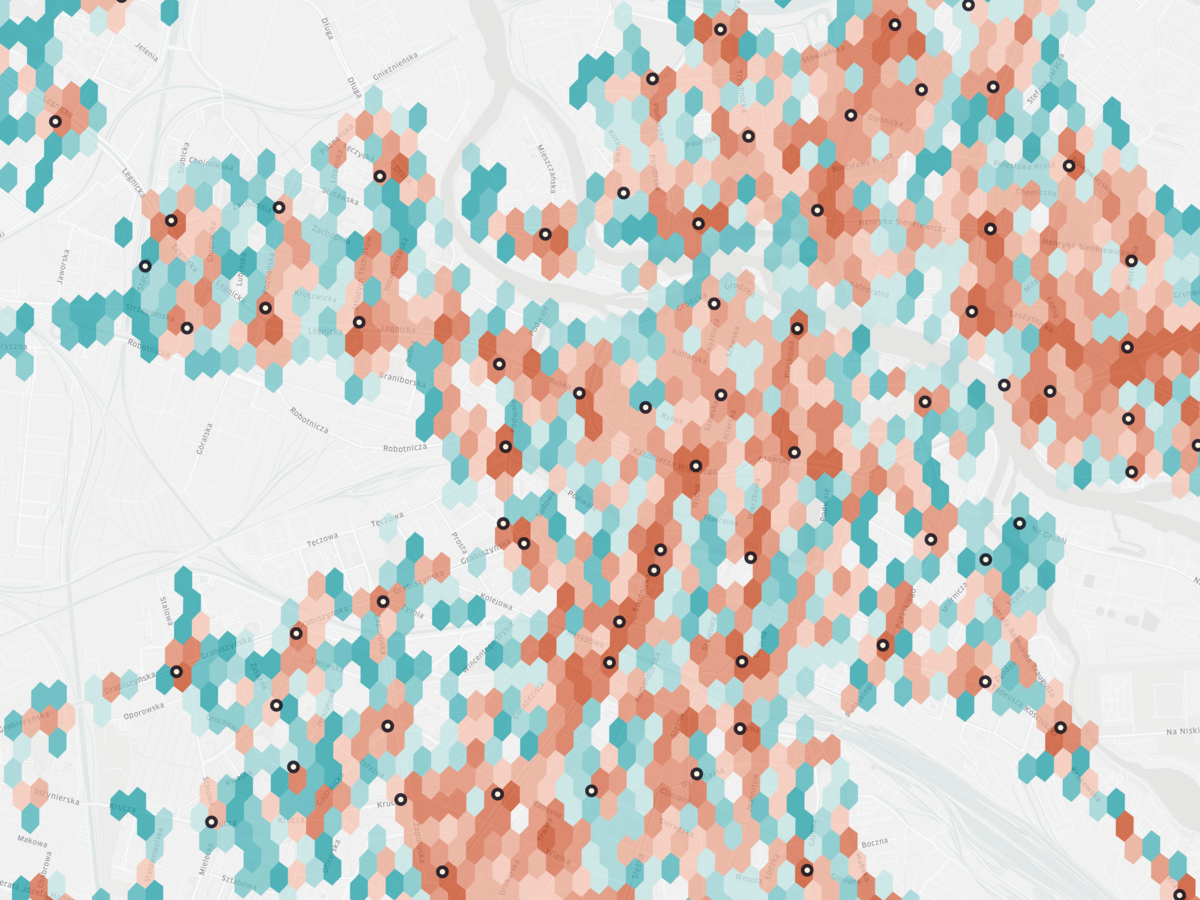}
    \caption{Method predictions for the city of Wroclaw for class imbalance ratio of 1.0.}
    \label{fig:wro_avg_imb:1}
\end{figure}

\begin{figure}[h]
    \centering
    \includegraphics[width=0.8\linewidth]{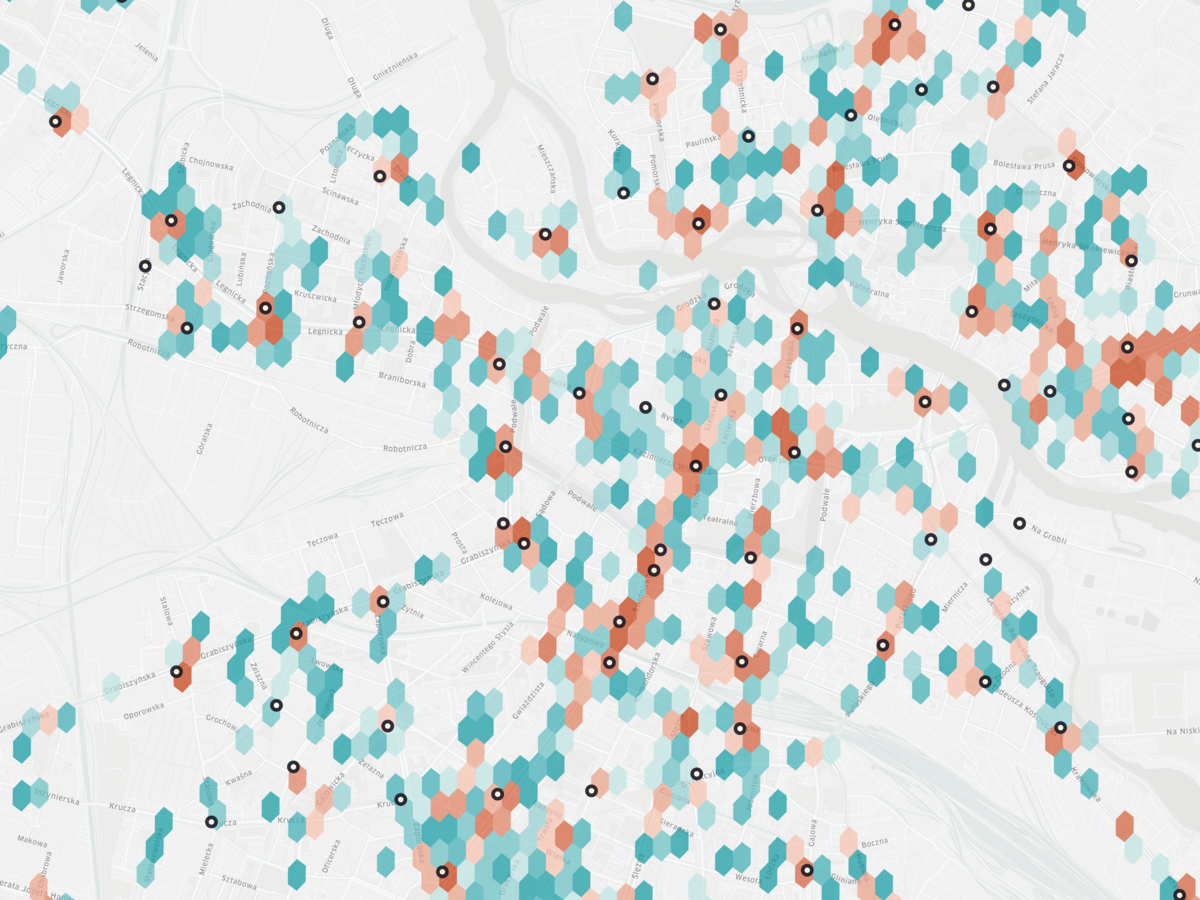}
    \caption{Method predictions for the city of Wroclaw for class imbalance ratio of 3.0.}
    \label{fig:wro_avg_imb:3}
\end{figure}

\begin{figure}[h]
    \centering
    \includegraphics[width=0.8\linewidth]{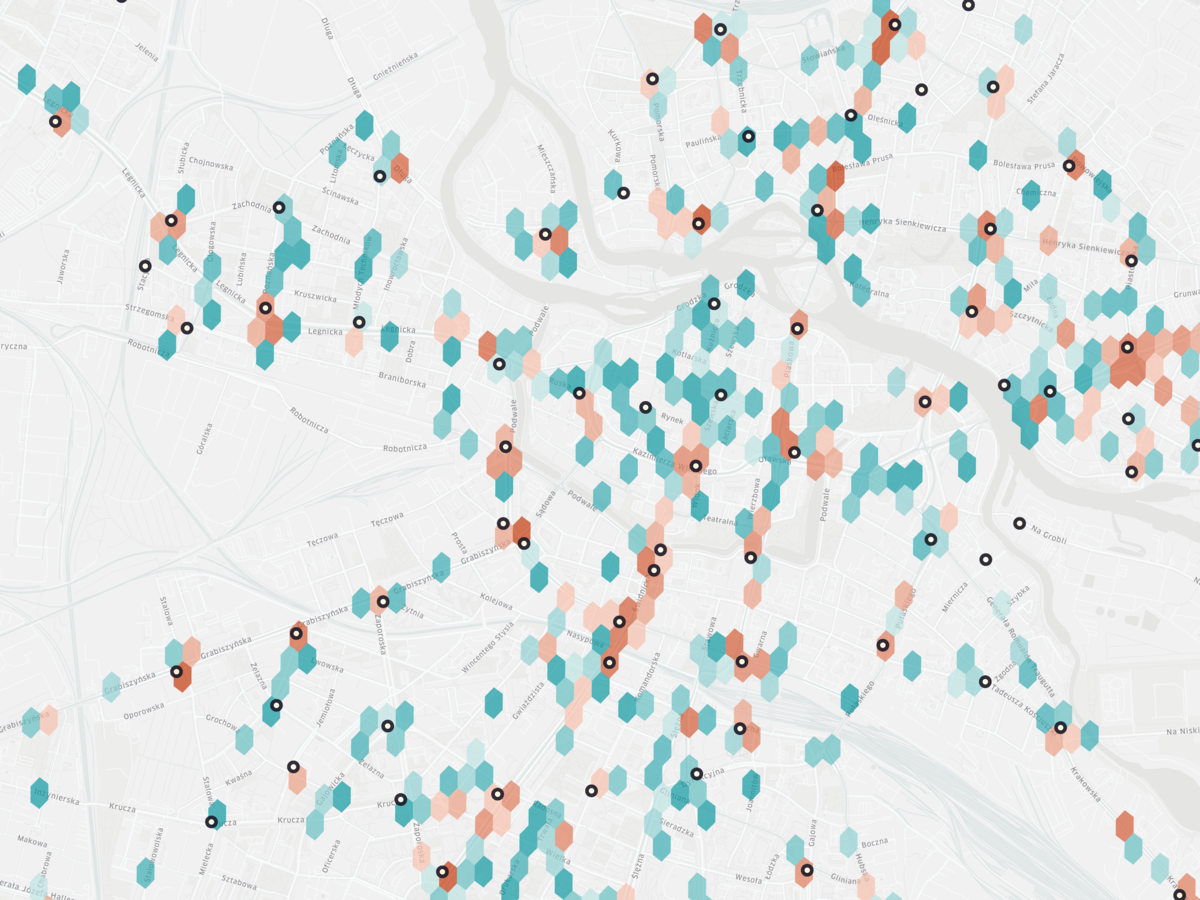}
    \caption{Method predictions for the city of Wroclaw for class imbalance ratio of 5.0.}
    \label{fig:wro_avg_imb:5}
\end{figure}

To better understand the influence of the hyperparameter on the performance quality, simulations for values 1, 2, 3, 4 and 5 were performed on the example of Wroclaw and visualised as heat maps. The simulations were run on resolution regions of 10 with a neighbourhood size of 10 and repeated 100 times and then averaged. The maps can be seen in Figures \ref{fig:wro_avg_imb:1}, \ref{fig:wro_avg_imb:3}, \ref{fig:wro_avg_imb:5}.

Based on the visualisation and precision and recall results, it was decided to choose a value for the class imbalance parameter equal to 2.5. It can be seen from the maps that for this value, not too many regions are marked as containing stations, which is a good sign because it is useful that the method does not mark the whole city as potential positions for stations. 

\subsection{Impact of the resolution and the size of region neighbourhood on the prediction quality}

\begin{figure*}[h]
    \centering
    \includegraphics[width=0.81\textwidth]{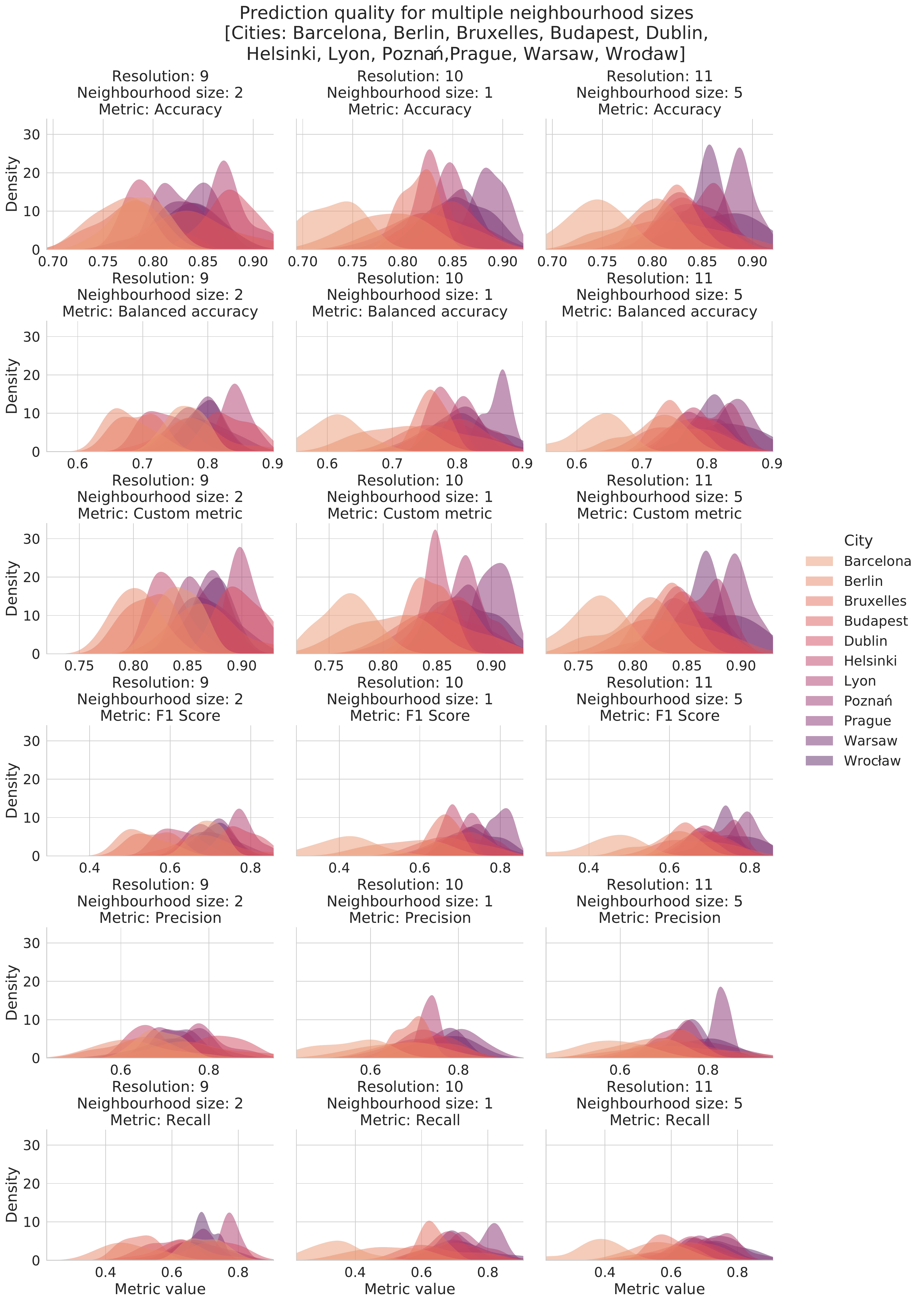}
    \caption[Prediction quality for multiple neighbourhood sizes]%
    {Prediction quality for multiple neighbourhood sizes. Results for 3 best candidates (one from each resolution).}
    \label{chart:comparing_neighbourhood_best}
\end{figure*}

From the point of view of the urban planner, the proposed method will be more useful, if it could more accurately indicate the region of the city in which the station should be located. Based on the previous results, it can be seen that as the resolution increases, the quality of the prediction decreases slightly. However, these were results averaged per resolution for many neighbourhood sizes. This question aims to find a suitable neighbourhood size that will produce good results but at the same time low enough not to unnecessarily increase the computational complexity of the method.
All results are summarised in the Table \ref{tab:comparing_neighbourhood}.
The results for the top 3 candidates 
are further detailed in Figure \ref{chart:comparing_neighbourhood_best}. 

As the results between the best candidates are similar, according to the decision to choose the highest resolution, value 11 was chosen.

The method predicts much more regions than only those that contain stations, which makes the method hard to evaluate using standard metrics for classification. However, it can be seen that the method manages to determine regions that contain stations with more certainty than their neighbourhood. As stations are not very densely distributed in cities, there may be situations where the model considers that a neighbourhood that does not actually contain a station should have one, based on the similarity of the urban significance of the neighbourhood regions.

\subsection{Method performance in station prediction task between cities}

After selection of all hyperparameters, the next step is to test the usability of the model in predicting stations in another city to be able to use the model learned on a city (or cities) that contains an existing bicycle sharing system and to propose a station layout on a city that does not have such a station layout. This will be tested by learning the model on a dataset from one city and trying to predict stations on a dataset from another city. However, a possible low quality of the prediction does not necessarily imply a weakness of the proposed method but could indicate that the city planners made different decisions about station layout and the station neighbourhoods are functionally completely different.

Using the final combination of hyperparameters, a cross-prediction of all 34 cities against each other individually was made. All hexes with stations were selected from the learning city and 2.5 times as many hexes without stations were drawn, which were then used to learn the model. The results were validated on the whole set of the second city. The experiments were repeated 100 times and averaged.

\begin{figure}[h]
    \centering
    \includegraphics[width=\linewidth]{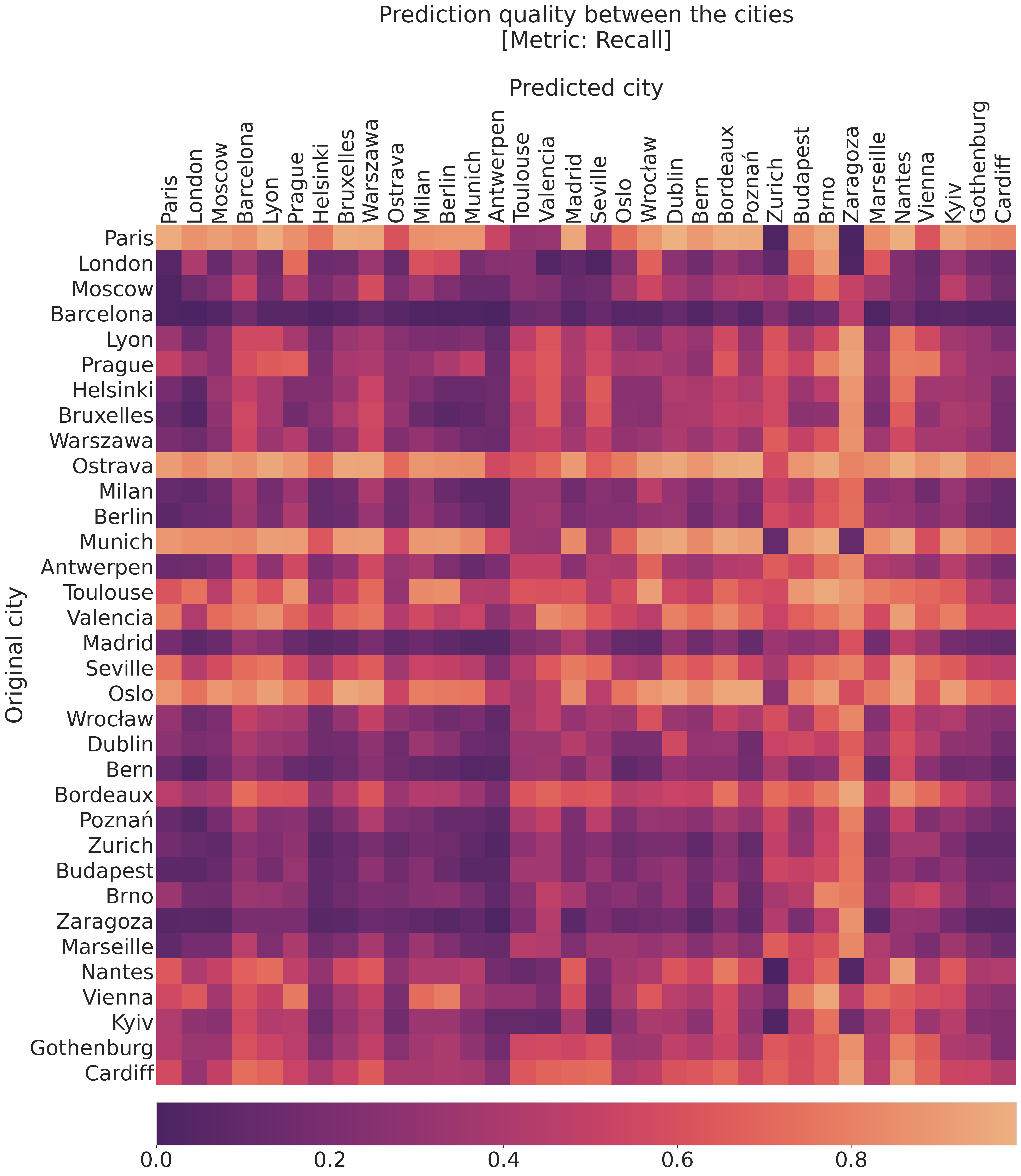}
    \caption[Recall metric in cross-city prediction]%
    {Recall metric in cross-city prediction. Cities were sorted in descending order of the number of stations in the city.}
    \label{chart:transfer_rec}
\end{figure}

Within the results, it was decided to present only the recall metric, as it is the only one that brings some information about the behavior of the model. Since the validation set included all hexes in the entire city, the ratio between hexes containing stations to those that did not contain stations was often 100 to 50-100 thousand. Therefore, the value of the accuracy metric was mostly above 98\%. The precision in most cases was very close to zero because the model predicted a lot of hexes that do not have stations as hexes that should contain a station (often proximity to stations). Along with the low precision, the F1 score was equally low. In the case of the recall metric, more correlations can be seen. The value of this measure varies between 0 and 0.95, and the heat map shows that some cities perform very well as base cities for station occurrence prediction (Munich, Oslo, Ostrava, Paris) and some cities, it is very difficult for the model to pick out all existing stations (Antwerp, Helsinki, Ostrava). Results are presented in the Figure \ref{chart:transfer_rec}.

The high value of the recall measure for some cities indicates that for those 4 best performing cities the model can predict well the occurrence of all stations within a city, and a low measure of accuracy (and precision) indicates that the model predicts many more positions as proposed than there are actually stations present, which is quite an expected result since cities often have multiple regions (especially at such high resolution) with similar structures that the model indicates slightly larger regions than a single hexagon. 

\begin{figure}[h]
    \centering
    \includegraphics[width=0.8\linewidth]{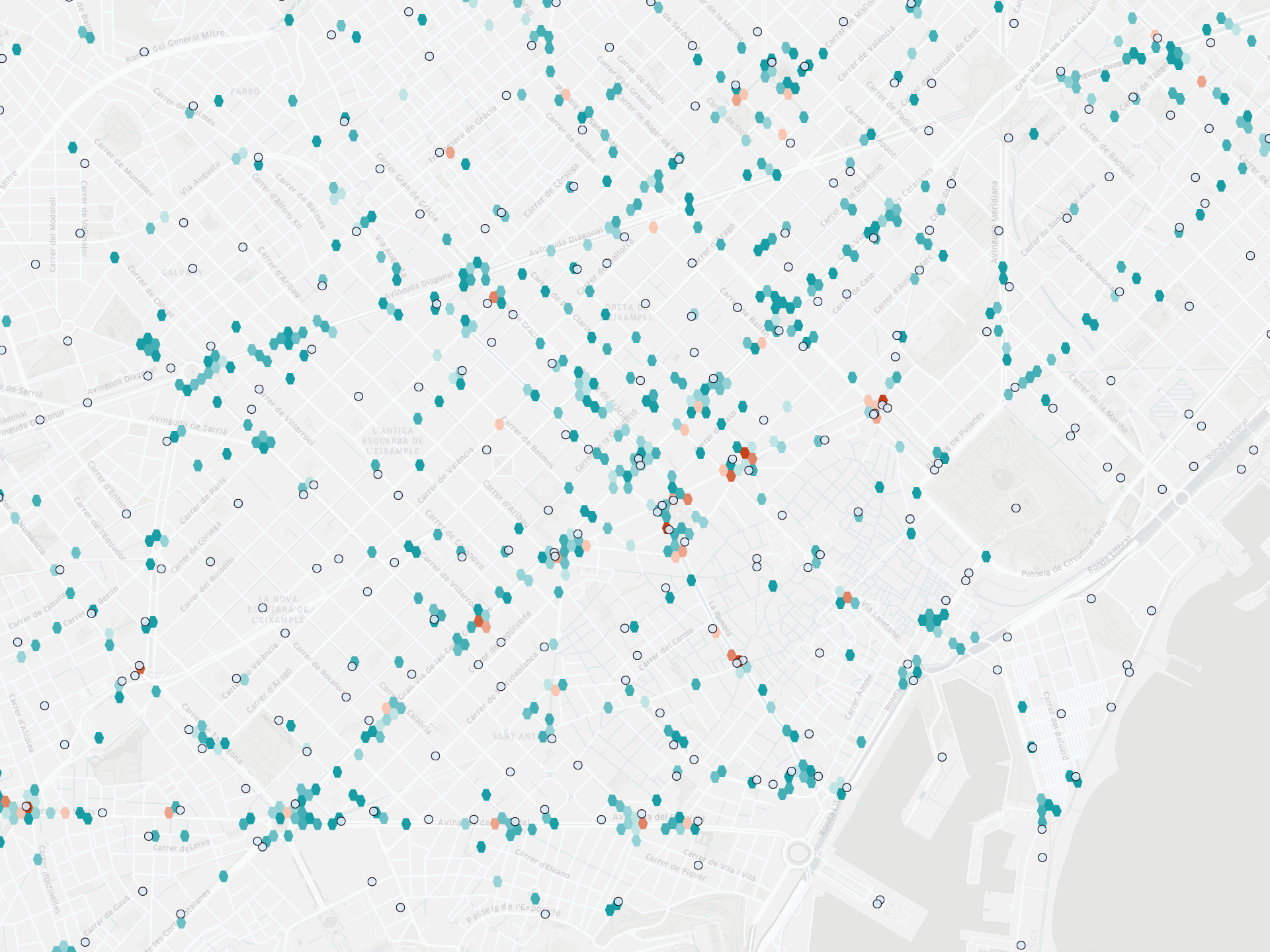}
    \caption[Example of Barcelona city predictions for method trained on a dataset from Barcelona]%
    {Example of Barcelona city predictions for method trained on a dataset from Barcelona. Results have been averaged from 100 iterations and filtered below probability 0.5. The dark teal colour represents the probability of 50\% and the dark red colour represents the probability of around 75\%. White dots with black borders represent existing stations.}
    \label{fig:barcelona_ex}
\end{figure}

Finally, it was decided to investigate how the predictions were distributed for the city of Barcelona, as the results indicated a very low recall measure for this city. The figure \ref{fig:barcelona_ex} shows a close-up map of the city centre with the predictions marked. It can be clearly seen that the method is not able to determine all station positions when the confidence threshold is equal to 0.5, which is the default value in the classifier. However, it cannot be said that the model generates completely random predictions and can reflect partial station layout infrastructure or predicts the occurrence of a station right next to where it occurs and because of the binary approach, the classifier is penalised for this result and the quality metric drops. This may be due to the density of the city, which in combination with the good quality tagging in OpenStreetMap causes many regions to generate embedding vectors that are close to each other in space. Additionally, it can be seen that the model performs better in the inner city than on the eastern coast of the city, for example.

\subsection{Example of predictions for a city without bicycle-sharing system}


To demonstrate the usefulness of the method, the results are presented below in the form of heat maps distributed over a city without bicycle sharing system - Florence (Italy) with over 350,000 inhabitants.

The 4 cities that had the best recall metric were used to teach the model: Munich, Oslo, Ostrava, Paris. The results are presented as 4 separate examples to show the differences in the performance of the model depending on the city chosen, which may allow other types of predictions to be made, which may depend on the layout of the city, the quality of its OpenStreetMap tagging and the layout of the bicycle sharing stations themselves. The examples are left without any judgement.

\begin{figure}[H]
    \centering
    \begin{minipage}{.25\textwidth}
        \centering
        \subfloat[Munich, Germany]{\label{fig:florence:38}\includegraphics[width=0.95\linewidth, keepaspectratio]{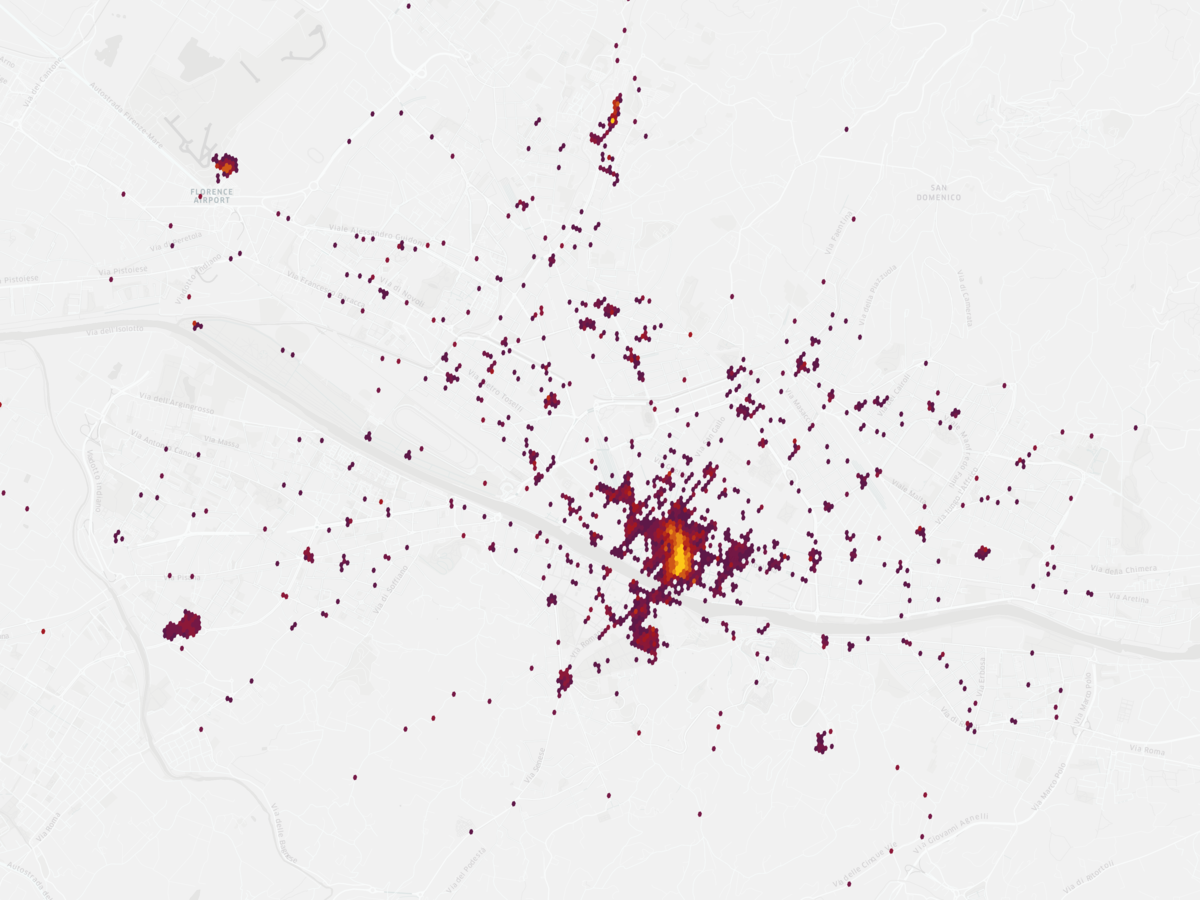}}
    \end{minipage}%
    \begin{minipage}{.25\textwidth}
        \centering
        \subfloat[Oslo, Norway]{\label{fig:florence:41}\includegraphics[width=0.95\linewidth, keepaspectratio]{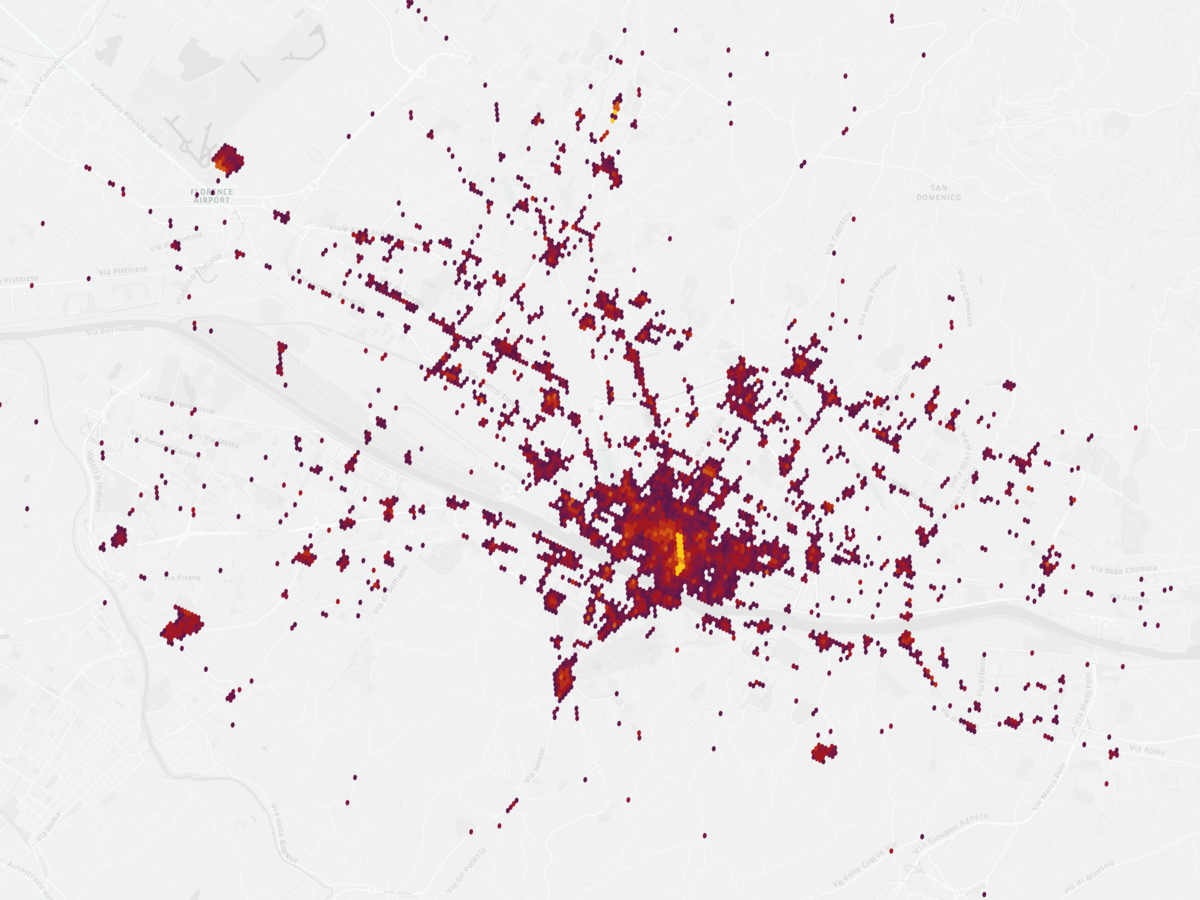}}
    \end{minipage}
    \par
    \begin{minipage}{.25\textwidth}
        \centering
        \subfloat[Ostrava, Czech Republic]{\label{fig:florence:42}\includegraphics[width=0.95\linewidth, keepaspectratio]{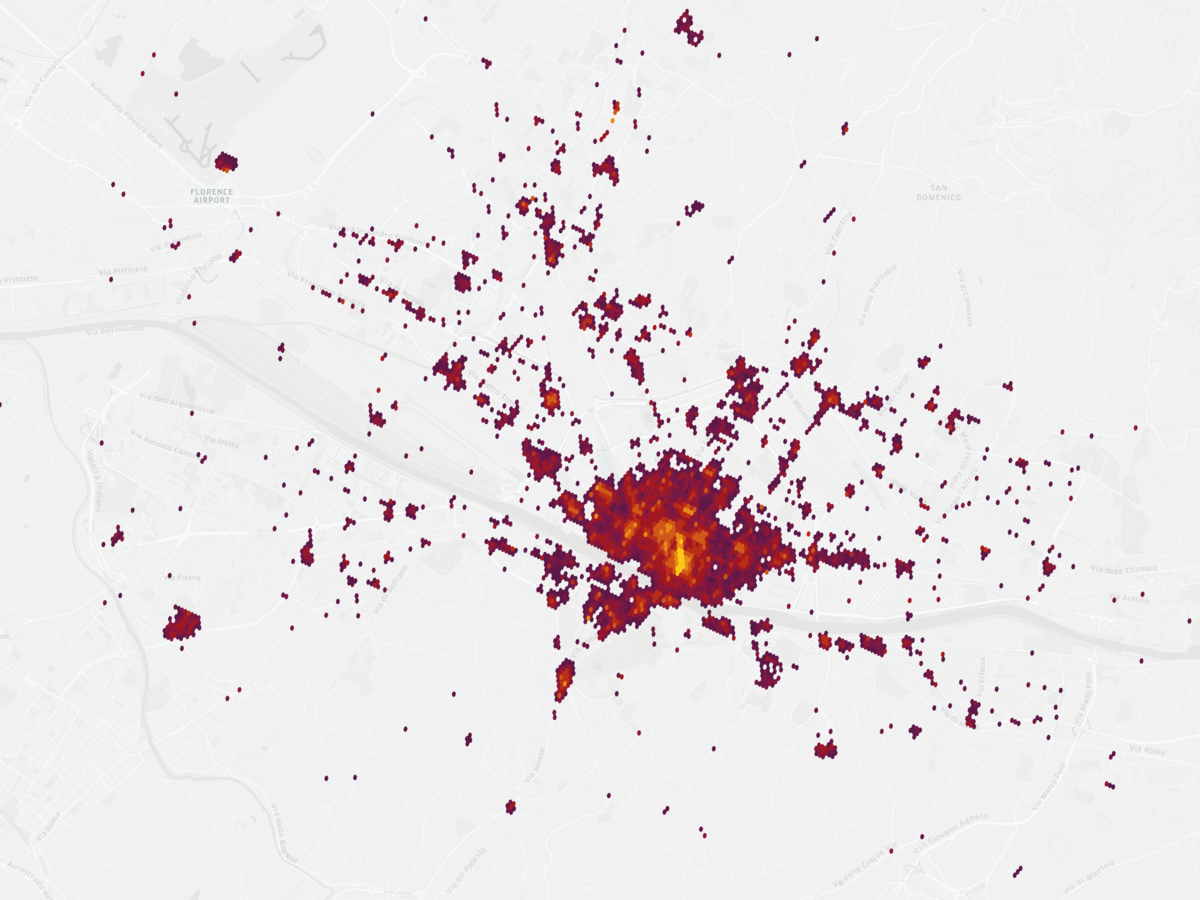}}
    \end{minipage}%
    \begin{minipage}{.25\textwidth}
        \centering
        \subfloat[Paris, France]{\label{fig:florence:45}\includegraphics[width=0.95\linewidth, keepaspectratio]{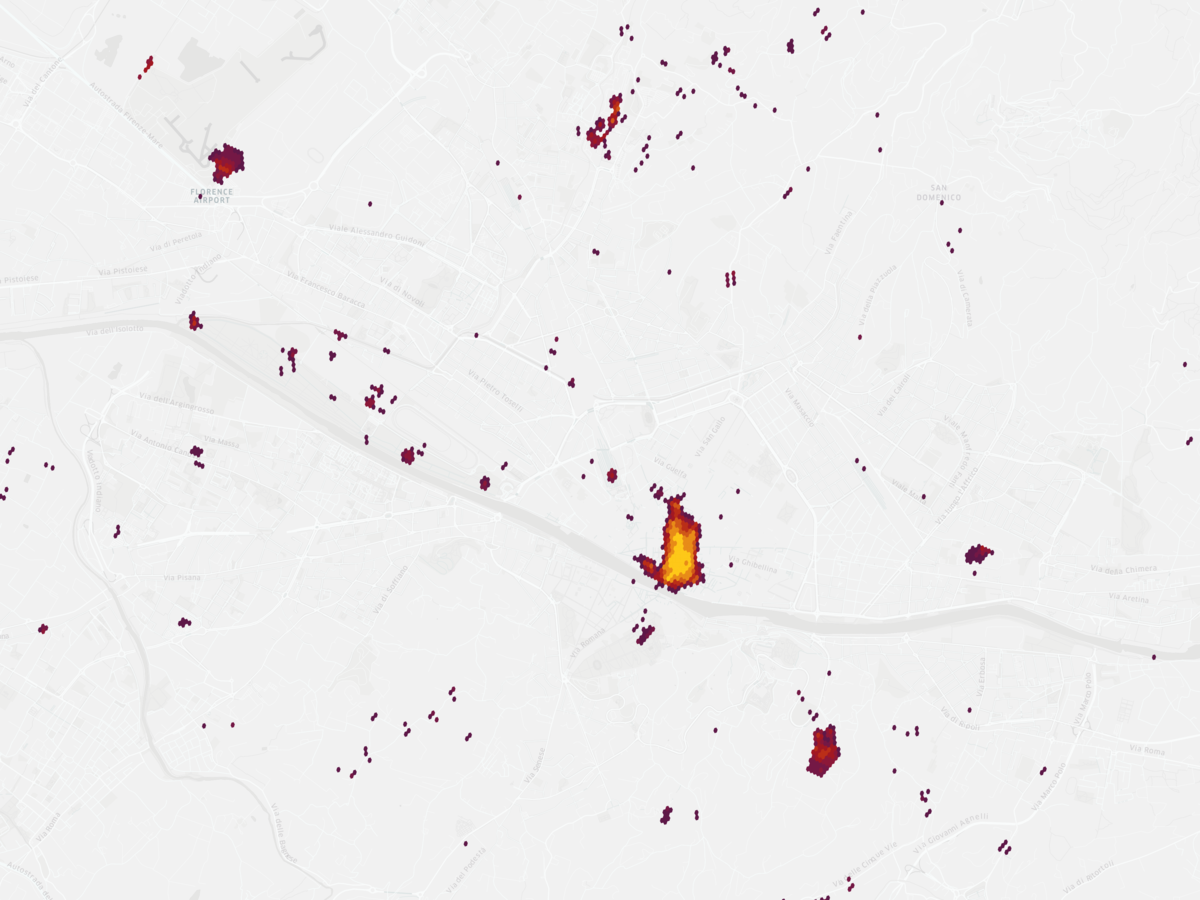}}
    \end{minipage}
    
    
    \caption[Example predictions for the city of Florence, Italy]%
    {Example predictions for the city of Florence, Italy.}
    \label{fig:florence}
\end{figure}

Results were averaged from 100 iterations. Each map represents only a selected subset of hexes above a selected threshold that varies from example to example because sometimes the model predicted (i.e., has a probability higher than 0.5) less than 50 hexes in the whole city. The threshold value varies between 0.2 and 0.5. Red colour represents low and yellow colour represents a high value of the predicted probability of occurrence of a station in a given microregion.

\section{Discussion}

\subsection{Summary of the results}

\paragraph*{Region neighbourhood embedding method.} \ \\*

Analysis of methods for combining neighbourhood vectors showed that a simple concatenation method performs quite well regardless of the neighbourhood size. Unfortunately, this method of concatenating vectors negatively affects the model learning performance as the dimensionality of the features increases significantly. The proposed averaging method, which is quite popular in natural language processing where the task vector is built by averaging vectors of single words, appeared to work well for low neighbourhood values, but with each successive ring of neighbours, the quality of prediction decreased. In addition to simple averaging, 2 weighted averaging methods were investigated: diminishing and squared diminishing. The results show that the last method behaves the most stable and obtains results similar to concatenation, so it was chosen as the method for neighbourhood embedding.

\paragraph*{Region embedding method.} \ \\*

A comparison of the different proposed methods for embedding single regions yielded unexpected results. It turned out that the basic method based only on the basic 20 categories without paying attention to the shapes of the studied objects works best. The method analyzing the shapes and taking into account the lengths of the roads in the regions gave similar results to the baseline method, but not higher values, so it was decided to stay with the first method. The more elaborate methods focusing on a higher number of tags and dimensionality reduction using an autoencoder were found to perform worse, however, this could be due to both too much granularity and sparseness of objects and differences in quality of tagging between cities, poor autoencoder infrastructure or poor architecture of the final NN classifier.

\paragraph*{Class imbalance ratio.} \ \\*

Examination of the class imbalance ratio showed that as it increased, the quality of both the F1 score and precision and recall measures decreased, which was a worrying sign. This could be because previously the model was taught on a subset but also tested on a subset rather than the whole city, and as the elements in the test set increased, the quality of the method performance decreased. Therefore, it was decided to visualise the results obtained by the method on the example of one of the cities. Based on the author's own assessment supported by the obtained heat maps and to maintain the average value of the F1 score measured higher than 0.6, it was decided to choose the ratio equal to 2.5. Thanks to it, the model was able to predict most of the existing stations, but on the other hand, it predicted a very large number of regions that do not have stations. However, showing a probability value of an occurrence rather than a binary score allows some regions to be filtered out by the user.

\paragraph*{Resolution and the size of region neighbourhood.} \ \\*

The final important element was the choice of the appropriate resolution and size of the neighbourhood considered for prediction. The analysis showed that all 3 resolutions with appropriate neighbourhood size were able to obtain similar results. As the aim of the prediction is to indicate the appropriate region as accurately as possible, it was decided to choose the highest resolution. It can be noticed that the differences between successive values of neighbourhoods are almost imperceptible, which may result from the chosen method of neighbourhood embedding. The difference between the vectors embedding the region with neighbourhood sizes 24 and 25 will be at most 1/125 or 0.008, which may be a small difference overall.

\paragraph*{Prediction between cities.} \ \\*

The previous research steps aimed to find the best set of hyperparameters allowing the method to reflect the station structure for the same city for which the classifier was learned. In this analysis, predictions were made between all pairs of cities to obtain information on whether the method can make valuable predictions between different cities.  The results indicate that the method obtains different results depending on the city, both the one used for learning and the one predicted. Going back to the exploratory analysis, it is difficult to establish any relationship between the calculated average population per station and the quality of prediction between cities that are close to each other in this space. However, one can see a certain correlation related to the average number of POIs per region - the cities of Bordeaux, Cardiff, Gothenburg, Munich, Oslo, Ostrava, Poznań, Seville, and Valencia, which have relatively few OSMs within their regions, allowing the method to obtain recall values above 0.8. On the other hand, Barcelona and Madrid, which are quite dense compared to other cities, caused the method to perform poorly in the presented binary station prediction task.
There are also a few city pairs that perform exceptionally poorly compared to other pairs in a row or column, namely Paris-Zurich, Paris-Zaragoza, Munich-Zurich, Munich-Zaragoza, Nantes-Zurich, Nantes-Zaragoza, Kyiv-Zurich, Kyiv-Zaragoza. Looking at the differences between these cities, it can be seen that for each of the pairs there are significant differences between the different characteristics. In some cities there is a significant difference in educational buildings, others in historical or leisure facilities, which are quite rare in Zaragoza and Zurich. However, it is difficult to judge whether this is the reason for such differences or whether it is due to another reason.

Due to the very poor recall that occurred for the city of Barcelona, it was decided to test how the method performs in predicting regions for this city. It turned out that the method does indeed omit existing stations or predicts the presence of a station right next to the place where the station actually occurs, which means that the method can learn from the significant features of the regions and what causes the station to be placed where it is, but it is possible that due to the dense city centre, the method averages the positions and the method is not able to perfectly reflect the existing station layout. 

\subsection{Hyperparameters selection}

Based on the experiments performed, the following list of hyperparameter values was selected:
\begin{itemize}
    \item Baseline classifier - Random forest
    \item Neighbourhood embedding method - Diminishing averaging squared
    \item Region embedding method - Category counting
    \item Features preprocessing - Normalisation (Min-Max)
    \item Imbalance ratio - 2.5
    \item Resolution - 11
    \item Neighbourhood size - 5
\end{itemize}

It is possible that for some cities it is worth using a different set of hyperparameters and the one chosen in the previous steps is not necessarily universal. It is possible that better results would be obtained in a linear regression task where each region would be assigned a value corresponding to the distance to the station (1 for the region with the station and decreasing for subsequent neighbourhoods). Another unexplored hyperparameter is the threshold level and its effect on prediction quality. Examining the prediction and recall curve would allow a more accurate diagnosis of the method's performance.

\section{Impact on resiliency}

As the scientific community published more and more reports from different cities around the globe \cite{campisi2020impact,hu2021systematic,teixeira2020link, habib2021impacts, teixeira2021motivations, schweizer2021outdoor} report, the coronavirus pandemic cause various disruptive shifts in urban mobility patterns. Urban green areas witness a rise in recreational biking during lockdowns or mobility restrictions. The initially unknown infection risk in public transport caused an  increase in bicycle usage for travel. Users who used bicycle sharing systems for the last mile part of their itineraries moved to use them for their entire trips. In areas without bicycle sharing systems, a rise of personal bike usage has been witnessed. Bicycles can be seen as the most affordable and safe mode of transport during extreme events. Providing citizens with a sufficient base of public bikes available in well distributed stations can be an important part of a resilient answer to a changing mobility behavior. However, developing a BSS is a complicated process which often requires expensive data gathering followed by mobility modelling, and can take many months to finish.

The method and models presented in this work can be used to shorten the response time and quickly provide city authorities with a hexagon based heatmap of how likely the hexagon is to be a good bike station locations. Such areas can be further processed manually by thresholding and human evaluation. They can also be clustered agglomeratively with a maximum distance limit set for points inside one cluster (ex. 150 meters) - such clusters can be used as candidate areas that undergo further investigation. Usually placing a bike stations requires additional checks such as enough pavement space or property ownership verification, working with a set of small areas will speed this process greatly. As a result city planners or authorities, and even citizens can be presented with a bike sharing system proposal in a matter of days and not months - strongly improving the ability of delivering a resilient response.

\section{Conclusion and future research}

Proposed in this research method can make inferences based only on OpenStreetMap data and return valuable information regarding the importance of different regions in the bicycle-sharing station layouts. The prediction for the whole city provides an easy-to-read analysis that can help planners to quickly identify areas that can be taken into account when designing station layouts. Both the function generating embedding vectors for the regions and the function converting these vectors into station probabilities have been thoroughly tested with different hyperparameters and allow valuable results to be obtained.

The method treats each region separately and returns station occurrence probabilities without paying attention to whether the region next to it has already been proposed as a station so that the method is not able to return a very accurate layout at this point but only a heat map that can assist decision-makers. However, this can be improved by adding another function that will select a user-specified number of stations together with a minimum distance between stations. An important value of this method is that it is easy to use - no nonpublic data was used for the analysis and it does not require any complex mobility analysis of the city. This research has focused entirely on the layout of bicycle stations, but nothing prevents it from being used for other tasks such as planning the layout of charging stations for electric vehicles. 

A natural development for further research is to extend the method to other continents and see how it will work on data from North America or Asia. For this work, it was decided to rely only on data from Europe to start the research with a small set. Another direction for the development of the method is to investigate what effect changing the algorithm from classification to regression would have. At first glance, the task seemed ideal to be solved by the binary classification method, however, the use of very high-resolution regions causes neighbouring regions to be close to each other in the embedding space and the method has problems drawing a definite line between the region where the station should be and where it should not be. Using regression and assigning different confidence values depending on, for example, the Euclidean distance or the distance calculated from the city's road layout, the method would be set to predict confidence from 0 to 1 from the start, instead of teaching the classification model and using the confidence returned by the classifier thanks to the implementation in the scikit-learn \cite{scikit-learn}  library.

Although the results can be easily interpreted by a person living in the city that the analysis concerns or by an expert dealing with urban planning topics, the added value would be to accurately determine the importance of the features on the decisions to be made to develop an explanatory model that can be better understood by the user. To increase the accuracy of the model, an algorithm could be added that takes into account how many stations should be distributed on the city grid and the minimum distances between stations to propose the exact regions that the stations should have, instead of returning a heat map of the city to the end-user.

\bibliographystyle{ACM-Reference-Format}
\bibliography{bibliography}

\newpage
\appendix
\section{Tabled Results}
\begin{table}[H]
\centering
\caption[Comparison of different region embedding methods]%
{Comparison of different region embedding methods. Results for the cities of Poznań, Warsaw, and Wrocław. Values represent the average of all grouped results for different parameters. Values in brackets represent the average score excluding the neighbourhood size of 0. Abbreviations: \textit{Acc.} - Accuracy, \textit{F1} - F1 Score, \textit{CC} - Category counting, \textit{SA} - Shape analysis per category, \textit{AT} - Shape analysis per all tags, \textit{ST} - Shape analysis per selected tags, \textit{RF} - Random forest classifier, \textit{NN} - Neural network classifier. The numbers next to \textit{NN} represent input embedding vector size. The best values in each row were highlighted.}
\label{tab:comparing_region_emb}
\resizebox{\linewidth}{!}{%
\begin{tabular}{@{}cc|ccccccc@{}}
\toprule
\begin{tabular}[c]{@{}c@{}}Hex\\ res.\end{tabular} &
  Metric &
  \begin{tabular}[c]{@{}c@{}}CC\\ RF\end{tabular} &
  \begin{tabular}[c]{@{}c@{}}CC\\ NN 20\end{tabular} &
  \begin{tabular}[c]{@{}c@{}}SA\\ RF\end{tabular} &
  \begin{tabular}[c]{@{}c@{}}SA\\ NN 36\end{tabular} &
  \begin{tabular}[c]{@{}c@{}}AT\\ NN 300\end{tabular} &
  \begin{tabular}[c]{@{}c@{}}ST\\ NN 20\end{tabular} &
  \begin{tabular}[c]{@{}c@{}}ST\\ NN 32\end{tabular} \\ \midrule
\multirow{2}{*}{9} &
  Acc. &
  \begin{tabular}[c]{@{}c@{}}\textbf{0.829}\\ (\textbf{0.829})\end{tabular} &
  \begin{tabular}[c]{@{}c@{}}0.811\\ (0.809)\end{tabular} &
  \begin{tabular}[c]{@{}c@{}}0.824\\ (0.823)\end{tabular} &
  \begin{tabular}[c]{@{}c@{}}0.788\\ (0.789)\end{tabular} &
  \begin{tabular}[c]{@{}c@{}}0.801\\ (0.803)\end{tabular} &
  \begin{tabular}[c]{@{}c@{}}0.778\\ (0.779)\end{tabular} &
  \begin{tabular}[c]{@{}c@{}}0.783\\ (0.785)\end{tabular} \\
 &
  F1 &
  \begin{tabular}[c]{@{}c@{}}\textbf{0.834}\\ (\textbf{0.834})\end{tabular} &
  \begin{tabular}[c]{@{}c@{}}0.812\\ (0.810)\end{tabular} &
  \begin{tabular}[c]{@{}c@{}}0.828\\ (0.828)\end{tabular} &
  \begin{tabular}[c]{@{}c@{}}0.788\\ (0.790)\end{tabular} &
  \begin{tabular}[c]{@{}c@{}}0.804\\ (0.807)\end{tabular} &
  \begin{tabular}[c]{@{}c@{}}0.775\\ (0.777)\end{tabular} &
  \begin{tabular}[c]{@{}c@{}}0.785\\ (0.787)\end{tabular} \\ \midrule
\multirow{2}{*}{10} &
  Acc. &
  \begin{tabular}[c]{@{}c@{}}\textbf{0.835}\\ (0.835)\end{tabular} &
  \begin{tabular}[c]{@{}c@{}}0.827\\ (0.825)\end{tabular} &
  \begin{tabular}[c]{@{}c@{}}0.833\\ (\textbf{0.837})\end{tabular} &
  \begin{tabular}[c]{@{}c@{}}0.803\\ (0.808)\end{tabular} &
  \begin{tabular}[c]{@{}c@{}}0.784\\ (0.788)\end{tabular} &
  \begin{tabular}[c]{@{}c@{}}0.790\\ (0.790)\end{tabular} &
  \begin{tabular}[c]{@{}c@{}}0.790\\ (0.792)\end{tabular} \\
 &
  F1 &
  \begin{tabular}[c]{@{}c@{}}\textbf{0.839}\\ (0.840)\end{tabular} &
  \begin{tabular}[c]{@{}c@{}}0.827\\ (0.824)\end{tabular} &
  \begin{tabular}[c]{@{}c@{}}0.838\\ (\textbf{0.843})\end{tabular} &
  \begin{tabular}[c]{@{}c@{}}0.800\\ (0.807)\end{tabular} &
  \begin{tabular}[c]{@{}c@{}}0.759\\ (0.764)\end{tabular} &
  \begin{tabular}[c]{@{}c@{}}0.776\\ (0.777)\end{tabular} &
  \begin{tabular}[c]{@{}c@{}}0.776\\ (0.779)\end{tabular} \\ \midrule
\multirow{2}{*}{11} &
  Acc. &
  \begin{tabular}[c]{@{}c@{}}\textbf{0.827}\\ (\textbf{0.834})\end{tabular} &
  \begin{tabular}[c]{@{}c@{}}0.820\\ (0.825)\end{tabular} &
  \begin{tabular}[c]{@{}c@{}}0.816\\ (0.826)\end{tabular} &
  \begin{tabular}[c]{@{}c@{}}0.775\\ (0.795)\end{tabular} &
  \begin{tabular}[c]{@{}c@{}}0.534\\ (0.535)\end{tabular} &
  \begin{tabular}[c]{@{}c@{}}0.726\\ (0.742)\end{tabular} &
  \begin{tabular}[c]{@{}c@{}}0.705\\ (0.714)\end{tabular} \\
 &
  F1 &
  \begin{tabular}[c]{@{}c@{}}\textbf{0.830}\\ (\textbf{0.838})\end{tabular} &
  \begin{tabular}[c]{@{}c@{}}0.820\\ (0.825)\end{tabular} &
  \begin{tabular}[c]{@{}c@{}}0.820\\ (0.830)\end{tabular} &
  \begin{tabular}[c]{@{}c@{}}0.756\\ (0.791)\end{tabular} &
  \begin{tabular}[c]{@{}c@{}}0.456\\ (0.444)\end{tabular} &
  \begin{tabular}[c]{@{}c@{}}0.669\\ (0.701)\end{tabular} &
  \begin{tabular}[c]{@{}c@{}}0.639\\ (0.658)\end{tabular} \\ \bottomrule
\end{tabular}%
}


\medskip

\centering
\resizebox{\linewidth}{!}{%
\begin{tabular}{@{}cc|ccccccc@{}}
\toprule
\begin{tabular}[c]{@{}c@{}}Hex\\ res.\end{tabular} &
  Metric &
  \begin{tabular}[c]{@{}c@{}}ST\\ NN 64\end{tabular} &
  \begin{tabular}[c]{@{}c@{}}ST\\ NN 100\end{tabular} &
  \begin{tabular}[c]{@{}c@{}}ST\\ NN 128\end{tabular} &
  \begin{tabular}[c]{@{}c@{}}ST\\ NN 200\end{tabular} &
  \begin{tabular}[c]{@{}c@{}}ST\\ NN 256\end{tabular} &
  \begin{tabular}[c]{@{}c@{}}ST\\ NN 300\end{tabular} &
  \begin{tabular}[c]{@{}c@{}}ST\\ NN 500\end{tabular} \\ \midrule
\multirow{2}{*}{9} &
  Acc. &
  \begin{tabular}[c]{@{}c@{}}0.799\\ (0.801)\end{tabular} &
  \begin{tabular}[c]{@{}c@{}}0.790\\ (0.790)\end{tabular} &
  \begin{tabular}[c]{@{}c@{}}0.799\\ (0.804)\end{tabular} &
  \begin{tabular}[c]{@{}c@{}}0.799\\ (0.802)\end{tabular} &
  \begin{tabular}[c]{@{}c@{}}0.791\\ (0.792)\end{tabular} &
  \begin{tabular}[c]{@{}c@{}}0.795\\ (0.798)\end{tabular} &
  \begin{tabular}[c]{@{}c@{}}0.798\\ (0.799)\end{tabular} \\
 &
  F1 &
  \begin{tabular}[c]{@{}c@{}}0.800\\ (0.802)\end{tabular} &
  \begin{tabular}[c]{@{}c@{}}0.792\\ (0.792)\end{tabular} &
  \begin{tabular}[c]{@{}c@{}}0.801\\ (0.806)\end{tabular} &
  \begin{tabular}[c]{@{}c@{}}0.794\\ (0.795)\end{tabular} &
  \begin{tabular}[c]{@{}c@{}}0.787\\ (0.789)\end{tabular} &
  \begin{tabular}[c]{@{}c@{}}0.793\\ (0.797)\end{tabular} &
  \begin{tabular}[c]{@{}c@{}}0.791\\ (0.792)\end{tabular} \\ \midrule
\multirow{2}{*}{10} &
  Acc. &
  \begin{tabular}[c]{@{}c@{}}0.792\\ (0.792)\end{tabular} &
  \begin{tabular}[c]{@{}c@{}}0.786\\ (0.788)\end{tabular} &
  \begin{tabular}[c]{@{}c@{}}0.788\\ (0.792)\end{tabular} &
  \begin{tabular}[c]{@{}c@{}}0.789\\ (0.792)\end{tabular} &
  \begin{tabular}[c]{@{}c@{}}0.789\\ (0.789)\end{tabular} &
  \begin{tabular}[c]{@{}c@{}}0.793\\ (0.793)\end{tabular} &
  \begin{tabular}[c]{@{}c@{}}0.788\\ (0.791)\end{tabular} \\
 &
  F1 &
  \begin{tabular}[c]{@{}c@{}}0.779\\ (0.779)\end{tabular} &
  \begin{tabular}[c]{@{}c@{}}0.771\\ (0.773)\end{tabular} &
  \begin{tabular}[c]{@{}c@{}}0.774\\ (0.779)\end{tabular} &
  \begin{tabular}[c]{@{}c@{}}0.775\\ (0.780)\end{tabular} &
  \begin{tabular}[c]{@{}c@{}}0.776\\ (0.777)\end{tabular} &
  \begin{tabular}[c]{@{}c@{}}0.780\\ (0.781)\end{tabular} &
  \begin{tabular}[c]{@{}c@{}}0.776\\ (0.781)\end{tabular} \\ \midrule
\multirow{2}{*}{11} &
  Acc. &
  \begin{tabular}[c]{@{}c@{}}0.705\\ (0.716)\end{tabular} &
  \begin{tabular}[c]{@{}c@{}}0.718\\ (0.733)\end{tabular} &
  \begin{tabular}[c]{@{}c@{}}0.717\\ (0.732)\end{tabular} &
  \begin{tabular}[c]{@{}c@{}}0.717\\ (0.733)\end{tabular} &
  \begin{tabular}[c]{@{}c@{}}0.715\\ (0.732)\end{tabular} &
  \begin{tabular}[c]{@{}c@{}}0.708\\ (0.721)\end{tabular} &
  \begin{tabular}[c]{@{}c@{}}0.693\\ (0.708)\end{tabular} \\
 &
  F1 &
  \begin{tabular}[c]{@{}c@{}}0.639\\ (0.657)\end{tabular} &
  \begin{tabular}[c]{@{}c@{}}0.676\\ (0.698)\end{tabular} &
  \begin{tabular}[c]{@{}c@{}}0.679\\ (0.703)\end{tabular} &
  \begin{tabular}[c]{@{}c@{}}0.674\\ (0.701)\end{tabular} &
  \begin{tabular}[c]{@{}c@{}}0.674\\ (0.698)\end{tabular} &
  \begin{tabular}[c]{@{}c@{}}0.670\\ (0.694)\end{tabular} &
  \begin{tabular}[c]{@{}c@{}}0.651\\ (0.672)\end{tabular} \\ \bottomrule
\end{tabular}%
}
\end{table}

\begin{table}[H]
\centering
\caption[Comparison of different neighbourhood sizes]%
{Comparison of different neighbourhood sizes. Results for the cities of Barcelona, Berlin, Brussels, Budapest, Dublin, Helsinki, Lyon, Poznań, Prague, Warsaw and Wrocław. Values represent the average of all grouped results. One best candidate neighbour size for each resolution has been highlighted.}
\label{tab:comparing_neighbourhood}
\resizebox{\linewidth}{!}{%
\begin{tabular}{@{}cc|cccc@{}}
\toprule
\begin{tabular}[c]{@{}c@{}}Hex\\ res.\end{tabular} & \begin{tabular}[c]{@{}c@{}}Neigh.\\ size\end{tabular} & Accuracy & F1 Score & Precision & Recall \\ \midrule
\multirow{4}{*}{9}   & 0          & 0.818          & 0.673          & 0.706          & 0.649          \\
                     & 1          & 0.822          & 0.679          & 0.711          & 0.656          \\
                     & \textbf{2} & \textbf{0.823} & \textbf{0.679} & \textbf{0.720} & \textbf{0.649} \\
                     & 3          & 0.817          & 0.667          & 0.710          & 0.635          \\ \midrule
\multirow{11}{*}{10} & 0          & 0.814          & 0.661          & 0.685          & 0.644          \\
                     & \textbf{1} & \textbf{0.828} & \textbf{0.677} & \textbf{0.715} & \textbf{0.650} \\
                     & 2          & 0.825          & 0.670          & 0.710          & 0.643          \\
                     & 3          & 0.828          & 0.673          & 0.721          & 0.641          \\
                     & 4          & 0.820          & 0.660          & 0.704          & 0.631          \\
                     & 5          & 0.822          & 0.667          & 0.705          & 0.639          \\
                     & 6          & 0.825          & 0.667          & 0.715          & 0.635          \\
                     & 7          & 0.827          & 0.671          & 0.721          & 0.637          \\
                     & 8          & 0.826          & 0.670          & 0.717          & 0.638          \\
                     & 9          & 0.829          & 0.676          & 0.723          & 0.642          \\
                     & 10         & 0.824          & 0.669          & 0.710          & 0.639          \\ \midrule
\multirow{26}{*}{11} & 0          & 0.798          & 0.606          & 0.679          & 0.552          \\
                     & 1          & 0.825          & 0.677          & 0.706          & 0.655          \\
                     & 2          & 0.826          & 0.675          & 0.715          & 0.648          \\
                     & 3          & 0.828          & 0.678          & 0.720          & 0.647          \\
                     & 4          & 0.825          & 0.669          & 0.713          & 0.638          \\
                     & \textbf{5} & \textbf{0.831} & \textbf{0.681} & \textbf{0.728} & \textbf{0.646} \\
                     & 6          & 0.824          & 0.668          & 0.712          & 0.636          \\
                     & 7          & 0.827          & 0.671          & 0.723          & 0.633          \\
                     & 8          & 0.828          & 0.672          & 0.719          & 0.638          \\
                     & 9          & 0.825          & 0.665          & 0.723          & 0.624          \\
                     & 10         & 0.827          & 0.671          & 0.723          & 0.634          \\
                     & 11         & 0.826          & 0.671          & 0.720          & 0.635          \\
                     & 12         & 0.823          & 0.663          & 0.714          & 0.626          \\
                     & 13         & 0.826          & 0.667          & 0.722          & 0.627          \\
                     & 14         & 0.829          & 0.674          & 0.729          & 0.634          \\
                     & 15         & 0.821          & 0.657          & 0.712          & 0.618          \\
                     & 16         & 0.827          & 0.665          & 0.726          & 0.621          \\
                     & 17         & 0.829          & 0.669          & 0.732          & 0.625          \\
                     & 18         & 0.828          & 0.673          & 0.721          & 0.638          \\
                     & 19         & 0.826          & 0.667          & 0.723          & 0.625          \\
                     & 20         & 0.825          & 0.664          & 0.722          & 0.622          \\
                     & 21         & 0.824          & 0.666          & 0.714          & 0.631          \\
                     & 22         & 0.828          & 0.671          & 0.726          & 0.630          \\
                     & 23         & 0.825          & 0.668          & 0.718          & 0.630          \\
                     & 24         & 0.826          & 0.666          & 0.722          & 0.625          \\
                     & 25         & 0.826          & 0.669          & 0.720          & 0.631          \\ \bottomrule
\end{tabular}%
}
\end{table}

\end{document}